\documentclass[runningheads]{llncs}
\usepackage{graphicx}

\usepackage{tikz}
\usepackage{comment}
\usepackage{amsmath,amssymb} 
\usepackage{color}
\usepackage{makecell}
\usepackage{wrapfig}

\usepackage[accsupp]{axessibility}  

\usepackage[hidelinks]{hyperref}
\usepackage{graphicx}
\usepackage{subfigure}
\usepackage{arydshln}
\usepackage{algorithmic}
\usepackage{cleveref}
\usepackage[symbol]{footmisc}
\crefname{section}{§}{§§}
\Crefname{section}{§}{§§}
\renewcommand{\thefootnote}{\fnsymbol{footnote}}

\begin{document}
\pagestyle{headings}
\mainmatter
\def\ECCVSubNumber{3598}  

\title{Mixed-Precision Neural Network Quantization via Learned Layer-wise Importance}

\author{
Chen Tang \inst{1}$^*$ \and
Kai Ouyang \inst{1}$^*$ \and
Zhi Wang \inst{1,4}$^\dagger$ \and
Yifei Zhu\inst{2} \and
Wen Ji\inst{3,4} \and \\
Yaowei Wang\inst{4} \and
Wenwu Zhu\inst{1}
}

\authorrunning{C. Tang, K. Ouyang et al.}
\institute{Tsinghua University\\
\and
Shanghai Jiao Tong University \\
 \and
Institute of Computing Technology, Chinese Academy of Sciences
\and Peng Cheng Laboratory\\
}

\titlerunning{MPQ via Learned Layer-wise Importance}

\maketitle

\def\thefootnote{*}\footnotetext{Equal contribution: \{tc20, oyk20\}@mails.tsinghua.edu.cn}\def\thefootnote{$\dagger$}\footnotetext{Corresponding author: wangzhi@sz.tsinghua.edu.cn}
\def\thefootnote{\arabic{footnote}}

\begin{abstract}
The exponentially large discrete search space in mixed-precision quantization (MPQ) makes it hard to determine the optimal bit-width for each layer. 
Previous works usually resort to \emph{iterative search} methods on the training set, which consume hundreds or even thousands of GPU-hours. 
In this study, we reveal that some unique learnable parameters in quantization, namely the scale factors in the quantizer, can serve as importance indicators of a layer, reflecting the contribution of that layer to the final accuracy at certain bit-widths.
These importance indicators naturally perceive the numerical transformation during quantization-aware training, which can precisely provide quantization sensitivity metrics of layers.
However, a deep network always contains hundreds of such indicators, and training them one by one would lead to an excessive time cost.
To overcome this issue, we propose a joint training scheme that can obtain all indicators at once.
It considerably speeds up the indicators training process by parallelizing the original sequential training processes.
With these learned importance indicators, we formulate the MPQ search problem as a \emph{one-time} integer linear programming (ILP) problem. 
That avoids the \emph{iterative} search and significantly reduces search time without limiting the bit-width search space.
For example, MPQ search on ResNet18 with our indicators takes only 0.06 seconds, which improves time efficiency exponentially compared to iterative search methods.
Also, extensive experiments show our approach can achieve SOTA accuracy on ImageNet for far-ranging models with various constraints (\emph{e.g.,} BitOps, compress rate). \\
Code is available on \url{https://github.com/1hunters/LIMPQ}. 

\keywords{Mixed-Precision Quantization, Model Compression}
\end{abstract}

\section{Introduction}
Neural network quantization can effectively compress the size and runtime overhead of a network by reducing the bit-width of the network. 
Using an equal bit-width for the entire network, a.k.a, fixed-precision quantization, is sub-optimal because different layers typically exhibit different sensitivities to quantization \cite{wang2019haq,cai2020rethinking}. 
It forces the quantization-insensitive layers to work at the same bit-width as the quantization-sensitive ones, missing the opportunity further to reduce the average bit-width of the whole network.

Mixed-precision quantization has thus become the focus of network quantization research, with its finer-grained quantization by allowing different bit-widths for different layers.  
In this way, the quantization-insensitive layers can use much lower bit-widths than the quantization-sensitive layers, thus providing more flexible accuracy-efficiency trade-off adjustment than the fixed-precision quantization.  
Finer-grained quantization also means exponentially larger searching space to search from. 
Suppose we have an $L$-layers network, each layer has $n$ optional bit-widths for weights and activations, the resulting search space is $n^{2L}$. 

Most of the prior works are search-based. HAQ \cite{wang2019haq} and AutoQ \cite{lou2019autoq} utilize deep reinforcement learning (DRL) to search the bit-widths by modeling bit-width determination problem as a Markov Decision Process.  
However, due to the exploration-exploitation dilemma, most existing DRL-based methods require a significant amount of time to finish the search process.
DNAS \cite{wu2018mixed} and SPOS \cite{guo2020single} apply Neural Architecture Search (NAS) algorithms to achieve a differentiable search process. 
As a common drawback of NAS, the search space needs to be greatly and manually limited in order to make the search process feasible, otherwise the search time can be quite high. 
In a word, the search-based approach is very time-consuming due to the need to evaluate the searched policy on the training set for multiple rounds (\emph{e.g.,} 600 rounds in HAQ \cite{wang2019haq}).

Different from these search-based approaches, some studies aim to define some ``critics'' to judge the quantization sensitivity of the layer. 
HAWQ \cite{dong2019hawq} and HAWQ-v2 \cite{dong2019hawq2} employ second-order information (Hessian eigenvalue or trace) to measure the sensitivity of layers and leverage them to allocate bit-widths. 
MPQCO \cite{chen2021towards} proposes an efficient approach to compute the Hessian matrix and formulate a Multiple-Choice Knapsack Problem (MCKP) to determine the bit-widths assignment.
Although these approaches reduce the searching time as compared to the search-based methods, they have the following defects: \\
\textit{(1) Biased approximation.} HAWQ and HAWQv2 approximate the Hessian information on the \emph{full-precision} (unquantized) network to measure the relative sensitivity of layers. 
This leads to not only an approximation error in these measurements themselves, but more importantly, an inability to perceive the existence of quantization operations.
A full-precision model is a far cry from a quantized model. 
We argue that using the information from the full-precision model to determine the bit-widths assignment of the quantized model is seriously biased and results in a sub-optimal searched MPQ policy. \\
\textit{(2) Limited search space.} 
MPQCO approximates its objective function with second-order Taylor expansion. 
However, the inherent problem in its expansion makes it impossible to quantize the activations with mixed-precision, which significantly limits the search space. 
A limited search space means that a large number of potentially excellent MPQ policies cannot be accessed during searching, making it more likely to result in sub-optimal performance due to a large number of MPQ policies being abandoned.
Moreover, MPQCO needs to assign the bit-witdhs of activations manually, which requires expert involvement and leaves a considerable room for improving search efficiency.

To tackle these problems, we propose to allocate bit-widths for each layer according to the \emph{learned end-to-end importance indicators}. 
Specifically, we reveal that the learnable scale factors in each layer's quantization function (\emph{i.e.,} quantizer), initially used to adjust the quantization mappings in classic quantization-aware training (QAT) \cite{esser2019learned,jung2019learning}, can be used as the importance indicators to distinguish whether one layer is more quantization-sensitive than other layers or not. 
As we will discuss later, they can perceive the numerical error transfer process and capture layers' characteristics in the quantization process (\emph{i.e.,} rounding and clamping) during QAT, resulting in a significant difference in the value of quantization-sensitive and insensitive layers. 
Since these indicators are learned end-to-end in QAT, errors that might arise from the approximation-based methods are avoided.
Moreover, the detached two indicators of each layer for weights and activations allow us to explore the whole search space without limitation.

Besides, an $L$-layer network with $n$ optional bit-widths for each layer's weights and activations has $M=2 \times L \times n$ importance indicators. 
Separately training these $M$ indicators requires $M$ training processes, which is time-prohibitive for deep networks and large-scale datasets.
To overcome this bottleneck, we propose a joint scheme to parallelize these $M$ training processes in a once QAT. 
That considerably reduces the indicators training processes by $M\times$.

Then, based on these obtained layer-wise importance indicators, we transform the original iterative MPQ search problem into a one-time ILP-based mixed-precision search to determine bit-widths for each layer automatically.
For example, a sensitive layer (\emph{i.e.,} larger importance) will receive a higher bit-width than an insensitive (\emph{i.e.,} smaller importance) layer. 
By this means, the time-consuming iterative search is eliminated, since we no longer need to use training data during the search. 
A concise comparison of our method and existing works is shown in Table \ref{tab:overall_comparision}.

\begin{table}[t]
\setlength{\tabcolsep}{0.19mm}
\setlength{}{}
\caption{A comparison of our method and existing works. Iterative search avoiding can significantly reduce the MPQ policy search time. Unlimited search space can provide more potentially excellent MPQ policies. Quantization-aware search can avoid the biased approximation on the full-precision model. Fully automatic bit-width assignment can effectively save human efforts and also reduce the MPQ policy search time. $^*$: MPQCO only can provide the quantization-aware search for weights.}
\begin{tabular}{c|c|c|c|c|c|c}
\hline
\small
Method                               & AutoQ & DNAS & HAWQ & HAWQv2 & MPQCO       & Ours \\ \hline
Iterative search avoiding            & No    & No   & Yes  & Yes    & Yes         & Yes  \\ \hline
Unlimited search space       & Yes   & No   & Yes  & Yes    & No          & Yes  \\ \hline
Quantization-aware search            & Yes   & Yes  & No   & No     & Partial yes$^*$ & Yes  \\ \hline
Fully automatic bit-width assignment & Yes   & Yes  & No   & Yes    & No          & Yes  \\ \hline
\end{tabular}
\label{tab:overall_comparision}
\vspace{-0.5cm}
\end{table}

To summarize, our contributions are the following:
\begin{itemize}
    \item 
    We demonstrate that a small number of learnable parameters (\emph{i.e.,} the scale factors in the quantizer) can act as importance indicators, to reflect the relative contribution of layers to performance in quantization. 
    These indicators are learned end-to-end without performing time-consuming fine-tuning or approximating quantization-unaware second-order information.
    \item 
    We transform the original \emph{iterative} MPQ search problem into a \emph{one-time} ILP problem by leveraging the learned importance of each layer, increasing time efficiency exponentially without limiting the bit-widths search space. Especially, we achieve about 330$\times$ MPQ policy search speedup compared to AutoQ on ResNet50, while preventing 1.7\% top-1 accuracy drop.
    \item 
    Extensive experiments are conducted on a bunch of models to demonstrate the state-of-the-art results of our method. 
    The accuracy gap between full-precision and quantized model of ResNet50 is further narrowed to only 0.6\%, while the model size is reduced by 12.2$\times$.
\end{itemize}

\section{Related Work}
\subsection{Neural Network Quantization}
\subsubsection{Fixed-Precision Quantization}
Fixed-precision quantization \cite{cai2017deep,zhou2017incremental,zhou2016dorefa,baskin2021nice} focus on using the same bit-width for all (or most of) the layers. 
In particular, \cite{zhang2018lq} introduces a learnable quantizer, \cite{choi2018pact} uses the learnable upper bound for activations. \cite{esser2019learned,jung2019learning} proposes to use the learnable scale factor (or quantization intervals) instead of the hand-crafted one.

\subsubsection{Mixed-Precision Quantization}    
To achieve a better balance between accuracy and efficiency, many mixed-precision quantization methods which search the optimal bit-width for each layer are proposed. 

\textit{Search-Based Methods.} 
Search-based methods aim to sample the vast search space of choosing bit-width assignments more effectively and obtain higher performance with fewer evaluation times. 
\cite{wang2019haq} and \cite{lou2019autoq} exploit DRL to determine the bit-widths automatically at a layer and kernel level. 
After that, \cite{uhlich2019mixed} determines the bit-width by parametrizing the optimal quantizer with the step size and dynamic range.
Furthermore, \cite{habi2020hmq} repurposes the Gumbel-Softmax estimator into a smooth estimator of a pair of quantization parameters.
In addition, many NAS-based methods have emerged recently \cite{wu2018mixed,yu2020search,cai2020rethinking,guo2020single}. 
They usually organize the MPQ search problem as a directed acyclic graph (DAG) and make the problem solvable by standard optimization methods (\emph{e.g.,} stochastic gradient descent) through differentiable NAS-based algorithms.
DiffQ \cite{defossez2021differentiable} uses pseudo quantization noise to perform differentiable quantization and search accordingly.

\textit{Criterion-Based Methods.}
Different from exploration approaches,
\cite{dong2019hawq} leverages the Hessian eigenvalue to judge the sensitivity of layers, and manually selects the bit width accordingly.
\cite{dong2019hawq2} further measures the sensitivity by the Hessian trace, and allocates the bit-width based on Pareto frontier automatically.
Furthermore, \cite{chen2021towards} reformulates the problem as a MCKP and proposes a greedy search algorithm to solve it efficiently. 
The successful achievement of criterion-based methods is that they reduce search costs greatly, but causing a biased approximation or limited search space as we discussed above. 

\subsection{Indicator-Based Model Compression} \label{sec:indicator_based_compression}
Measuring the importance of layers or channels using learned (\emph{e.g.,} scaling factors of batch normalization layers) or approximated indicators are seen as promising work thanks to its excellent efficiency and performance.
Early pruning work \cite{lecun1990optimal} uses second-derivative information to make a trade-off between network complexity and accuracy.
\cite{liu2017learning} pruning the unimportant channels according to the corresponding BN layer scale factors.
\cite{chen2021bn} sums the scale factors of BN layer to decide which corresponding convolution layer to choose in NAS search process. 
However, quantization is inherently different from these studies due to the presence of numerical precision transformation.
\section{Method}
\subsection{Quantization Preliminary}
\label{sec:understand_factor}
Quantization maps the continuous values to discrete values. The uniform quantization function (a.k.a quantizer) under $b$ bits in QAT maps the input $float32$ activations and weights to the homologous quantized values $[0, 2^{b}-1]$ and $[-2^{b-1}, \\2^{b-1}-1]$. The quantization functions $Q_b(\cdot)$ that quantize the input values $v$ to quantized values $v^q$ can be expressed as follows:

\begin{equation} 
v^q=Q_b(v;s)= round(clip(\frac{v}{s},min_b,max_b)) \times s,
\label{eq:preliminary}
\end{equation}
where $min_b$ and $max_b$ are the minimum and maximum quantization value \cite{bhalgat2020lsq+,esser2019learned}. For activations, $min_b=0$ and $max_b=2^{b}-1$. For weights, $min_b=-2^{b-1}$ and $max_b=2^{b-1}-1$. 
$s$ is a learnable scalar parameter used to adjust the quantization mappings, called the \emph{step-size scale factor}. For a network, each layer has two distinct scale factors in the weights and activations quantizer, respectively. 

To understand the role of the scale factor, we consider a toy quantizer example under $b$ bits and omit the $clip(\cdot)$ function. 
Namely,
\begin{equation} 
v^q=round(\frac{v}{s}) \times s=\hat{v^q} \times s,
\label{eq:preliminary_example}
\end{equation}
where $\hat{v^q}$ is the quantized integer value on the discrete domain. 

Obviously, for two continuous values $v_i$ and $v_j$ ($v_i \neq v_j$), their quantized integer values $\left|\hat{v^q_i}-\hat{v^q_j}\right|=0$ if and only if $0 < \left|v_i-v_j\right| \leq \frac{s}{2}$. 
Thus $s$ actually controls the distance between two adjacent quantized values.
A larger $s$ means that more different continuous values are mapped to the same quantized value.

\subsection{From Accuracy to Layer-wise Importance}
Suppose we have an $L$-layer network with full-precision parameter tensor $\mathcal{W}$, each layer has $n$ optional bit-widths $\mathcal{B}=\{b_0, ..., b_{n-1}\}$ for activation and weights of each layer, respectively. 
The bit-width combination of weights and activations $(b^{(l)}_{w}, b^{(l)}_{a})$ for layer $l$ is $b^{(l)}_{w} \in \mathcal{B}$ and $b^{(l)}_{a} \in \mathcal{B}$. 
Thus $\mathcal{S}=\{(b^{(l)}_w,b^{(l)}_a)\}_{l=0}^L$ is the bit-width combination for the whole network, and we use $\mathcal{W_S}$ to denote the quantized parameter tensor. All possible $\mathcal{S}$ construct the search space $\mathcal{A}$.
Mixed-precision quantization aims to find the appropriate bit-width combination (\emph{i.e.,} searched MPQ policy) $\mathcal{S^*} \in \mathcal{A}$ for the whole network to maximize the validation accuracy $\mathcal{ACC}_{val}$, under certain constraints $C$ (\emph{e.g.,} model size, BitOps, etc.). 
The objective can be formalized as follows:

\vspace{-0.5cm}
\begin{subequations}\label{eq:mp_original}
\begin{align}
\mathcal{S^*}=\mathop{\arg\max}\limits_{\mathcal{S} \thicksim \Gamma(\mathcal{A})} \mathcal{ACC}_{val}(f(\textbf{x}; \mathcal{S}, \mathcal{W_\mathcal{S}}), \textbf{y})   \nonumber \ \ \nonumber \tag{\ref{eq:mp_original}}
\end{align} 
\vspace{-0.3cm}
\begin{alignat}{2}
\text{s.t.} \quad &  \mathcal{W_\mathcal{S}}=\mathop{\arg\min}_\mathcal{W}\mathcal{L}_{train}(f(\textbf{x}; \mathcal{S}, \mathcal{W}), \textbf{y}) \\
& \quad \quad \quad BitOps(\mathcal{S}) \leq C 
\end{alignat}
\end{subequations}
where $f(\cdot)$ denotes the network, $\mathcal{L}(\cdot)$ is the loss function of task (\emph{e.g.,} cross-entropy), \textbf{x} and \textbf{y} are the input data and labels, $\Gamma(\mathcal{A})$ is the prior distribution of $\mathcal{S} \in \mathcal{A}$. For simplicity, we omit the data symbol of training set and validation set, and the parameter tensor of quantizer. This optimization problem is combinatorial and intractable, since it has an extremely large discrete search space $\mathcal{A}$. As above, although it can be solvable by DRL or NAS methods, the time cost is still very expensive. This is due to the need to evaluate the goodness of a specific quantization policy $\mathcal{S}$ on the training set to obtain metrics $\mathcal{L}_{train}$ iteratively to guide the ensuing search. As an example, AutoQ \cite{lou2019autoq} needs more than 1000 GPU-hours to determine a final quantization strategy $\mathcal{S^*}$ \cite{chen2021towards}.

Therefore, we focus on \emph{replacing the iterative evaluation on the training set with some once-obtained importance score of each layer}. 
In this way, the layer-wise importance score indicates the impact of quantization between and within each layer on the final performance, thus avoiding time-consuming iterative accuracy evaluations. 
Unlike the approximated Hessian-based approach \cite{dong2019hawq,dong2019hawq2}, which is imperceptible to quantization operations or limits the search space \cite{chen2021towards}, we propose to \emph{learn the importance in the Quantization-Aware Training}.

\subsection{Learned Layer-wise Importance Indicators}
Quantization mapping is critical for a quantized layer since it decides how to use confined quantization levels, and improper mapping is harmful to the performance \cite{jung2019learning}.
As shown in Equation \ref{eq:preliminary}, during QAT, the learnable scale factor of the quantizer in each layer is trained to adjust the corresponding quantization mapping \emph{properly} at a specific bit-width. 
This means that it can naturally capture certain quantization characteristics to describe the layers due to its controlled quantization mapping being optimized directly by the task loss. 

\begin{figure}[h]
\centering
\includegraphics[scale=0.4]{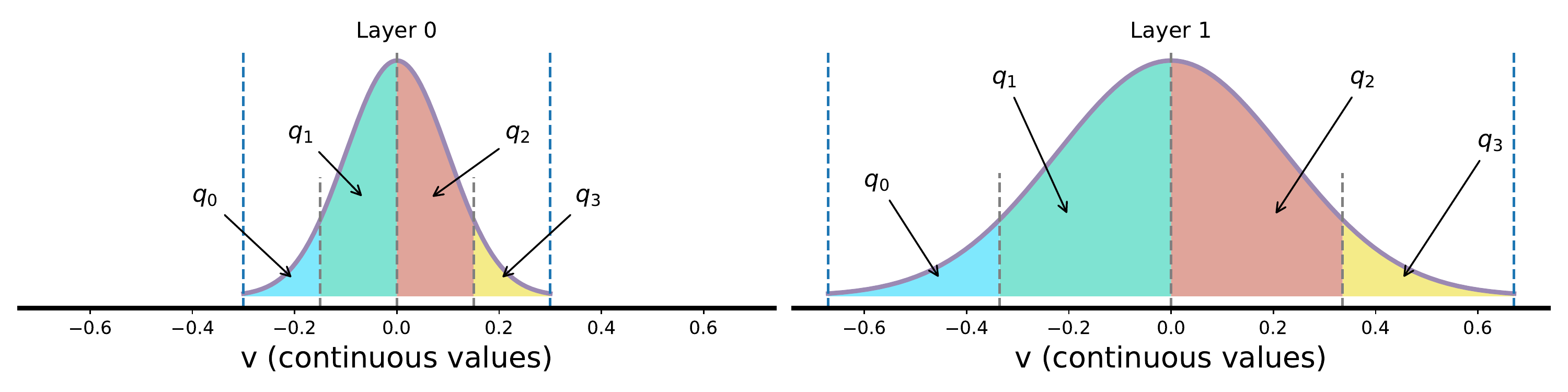}
\caption{Illustration of distribution for two example layers, both under 2 bits quantization. 
The grey dashed line (representing the scale factor of quantizer $s$) separates the different quantization levels (\emph{i.e.,} $2^2=4$ quantized values \{$q_0$, $q_1$, $q_2$, $q_3$\} for 2 bits).
For example, the continuous values in green and red area are quantized to the same quantization level $q_1$ and $q_2$, respectively.
}
\label{fig:motivation_vis}
\end{figure}

As we discussed in \cref{sec:understand_factor}, the continuous values in a uniform range fall into the same quantization level, the specific range is controlled by the scale factor $s$ of this layer.
We consider two example layers with \emph{well-trained} $s$ and weights (\emph{i.e.,}, both in a local minimum after a quantization-aware training) and quantized through Equation \ref{eq:preliminary_example}.
As shown in Figure. \ref{fig:motivation_vis}, the continuous distribution of layer 1 is much wider than layer 0.
According to the analysis in \cref{sec:understand_factor}, this results in layer 1 more different continuous values being mapped to the same quantized values (\emph{e.g.,} the values in green area are mapped to $q_1$), and therefore having a large learned scale factor $s$.
However, while the green area in layer 0 and layer 1 are both quantized to the value $q_1$, the green area of layer 1 contains a much broader continuous range, and the same goes for other areas.
In extreme cases, this extinguishes the inherent differences of original continuous values, thus reducing the expressiveness of the quantized model \cite{park2020profit}.
To overcome this and maintain the numerical diversity, we should give more available quantization levels to those layers with large scale factors, namely, increasing their bit-width.

Therefore, the numerically significant difference in the learned scale factors of heterogeneous layers can properly assist us to judge the sensitivity of layer.
Moreover, the operation involved in the scale factor takes place in the quantizer, which allows it to be directly aware of quantization. 
Last but not least, there are two quantizers for activations and weights for a layer, respectively, which means that we can obtain the importance of weights and activations separately. 

\subsubsection{Feasibility Verification}
Despite the success of indicator-based methods for model compression \cite{chen2021bn,lecun1990optimal,liu2017learning} to avoid a time-consuming search process, to the best of our knowledge, there is no literature to demonstrate that the end-to-end learned importance indicators can be used for quantization. To verify the scale factors of quantizer can be used for this purpose, we conduct a contrast experiment for MobileNetv1 \cite{howard2017mobilenets} on ImageNet \cite{deng2009imagenet} as follows. 

In the MobileNet family, it is well-known that the depth-wise convolutions (DW-convs) have fewer parameters than the point-wise convolutions (PW-convs); thus, the DW-convs are generally more susceptible to quantization than PW-convs \cite{habi2020hmq,park2020profit}. 
Therefore, we separately quantized the DW-conv and PW-conv for each of the five DW-PW pairs in MobileNetv1 to observe whether the scale factors of the quantizer and accuracy vary. 
Specifically, we quantized each layer in MobileNetv1 to 2 or 4 bits to observe the accuracy degradation. Each time we quantized only \emph{one layer} to low bits while other layers are not quantized and updated, \emph{i.e.,} we quantized 20 $(5 \times 2 \times 2)$ networks independently. If the accuracy of layer $l_i$ degrades higher when the quantization level changes from 4 bits to 2 bits than layer $l_j$, then $l_i$ is \emph{more sensitive} to quantization than $l_j$. In addition, the input channels and output channels of these five DW-PW pairs are all 512. Namely, we used the same number of I/O channels to control the variables. 
\begin{wrapfigure}[]{r}{0.65\textwidth}
\centering
\includegraphics[scale=0.45]{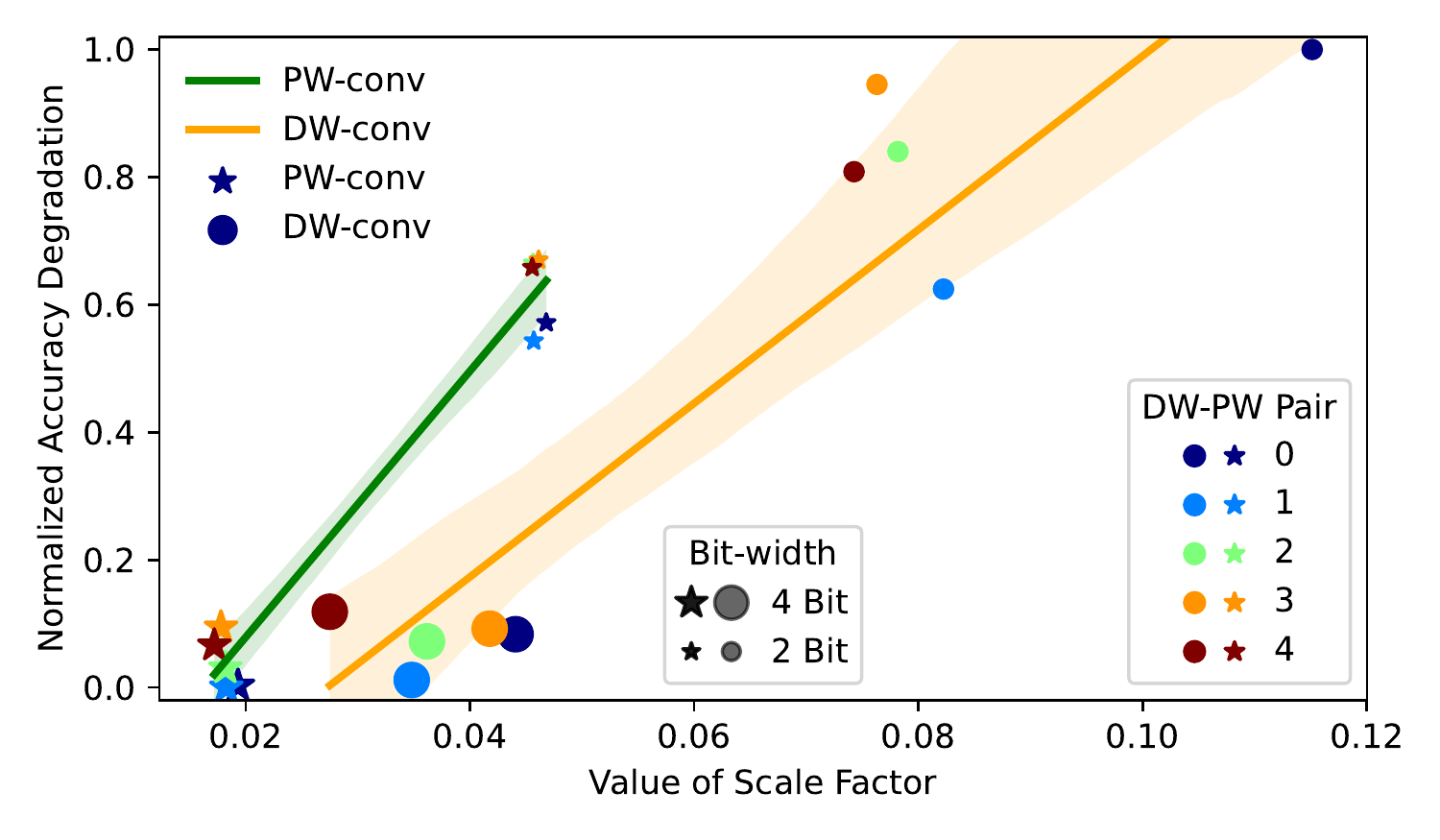}
\caption{Results of the contrast experiment of MobileNetv1 on ImageNet. ``$\bullet$'' and ``$\star$'' respectively indicate that the DW-conv layer or the PW-conv layer is quantized. Different colors indicate that different layers are quantized. Large labels indicate that the quantization bit-width is set to 4 bits and small labels of 2 bits.}
\label{fig:mbv1_separate_importance}
\end{wrapfigure}

The results of the controlled variable experiment are shown in Figure \ref{fig:mbv1_separate_importance}. 
Based on the results, we can draw the following conclusions: 
When the quantization bit-width decreases from 4 to 2 bits, the accuracy degradation of PW-convs is much lower than that of DW-convs, which consists of the prior knowledge that DW-convs are very sensitive.
Meanwhile, the values of scale factors of all PW-convs are prominent smaller than those of DW-convs under the same bit-width. 
That indicates the values of scale factor of whose sensitive layers are bigger than whose insensitive layers, which means the scale factor's value can adequately reflect the quantization sensitivity of the corresponding layer.
Namely, the kind of layer with a large scale factor value is more important than the one with a small scale factor. 

\subsubsection{Initialization of the Importance Indicators}
Initializing the scale factors with the statistics \cite{bhalgat2020lsq+,esser2019learned} of each layer results in the different initialization for each layer. 
We verify whether the factors still show numerical differences by the same initialization value scheme to erase this initialization difference. 
That is, for each layer, we empirically initialize each importance indicator of bit $b$ by $s_b = 0.1 \times \frac{1}{b}$ since we observed the value of factor is usually quite small ($\leq$ 0.1) and increases as the bit-width decreases.
\begin{figure}[!htb]
\begin{tabular}{cc}
\hspace{-0.3cm}
\begin{minipage}[t]{0.25\linewidth}
    \includegraphics[width = 1\linewidth]{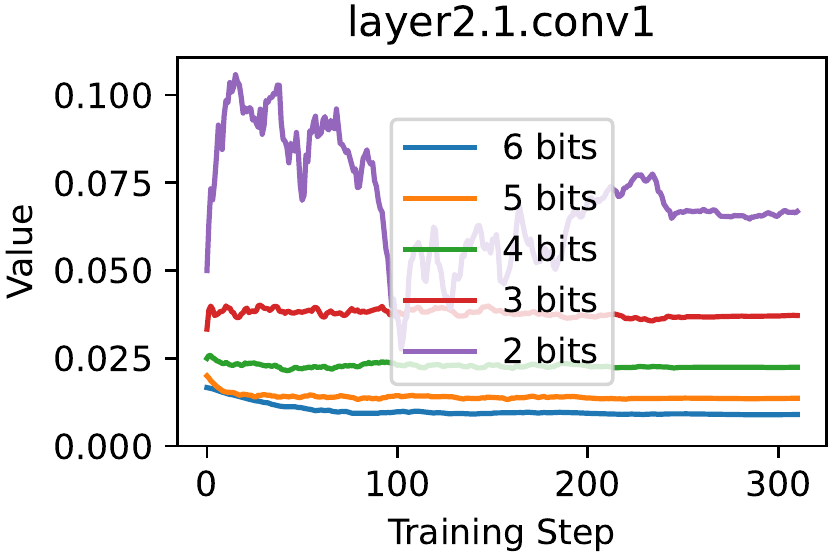}
\end{minipage}
\hspace{-0.2cm}
\begin{minipage}[t]{0.25\linewidth}
    \includegraphics[width = 1\linewidth]{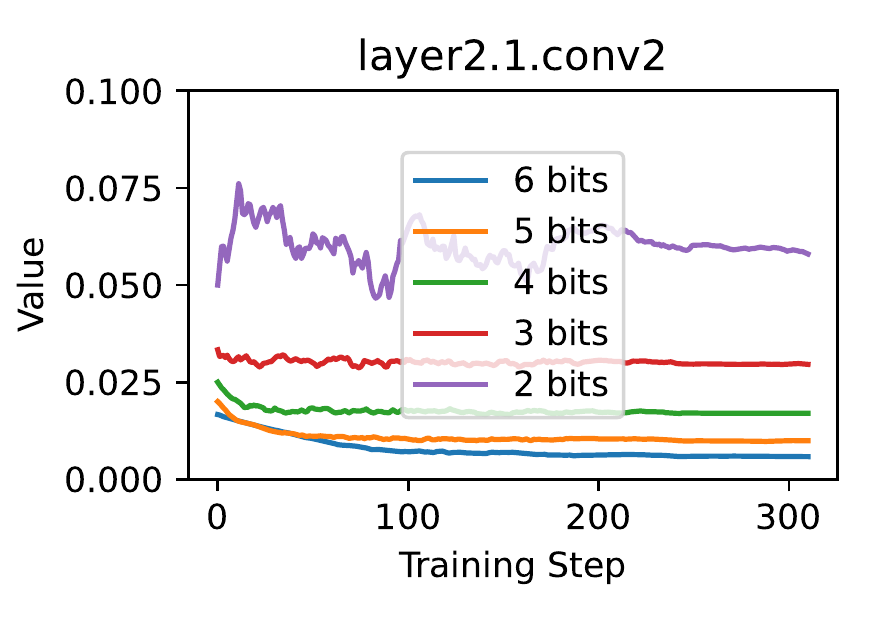}
\end{minipage}
\hspace{-0.2cm}
\begin{minipage}[t]{0.25\linewidth}
    \includegraphics[width = 1\linewidth]{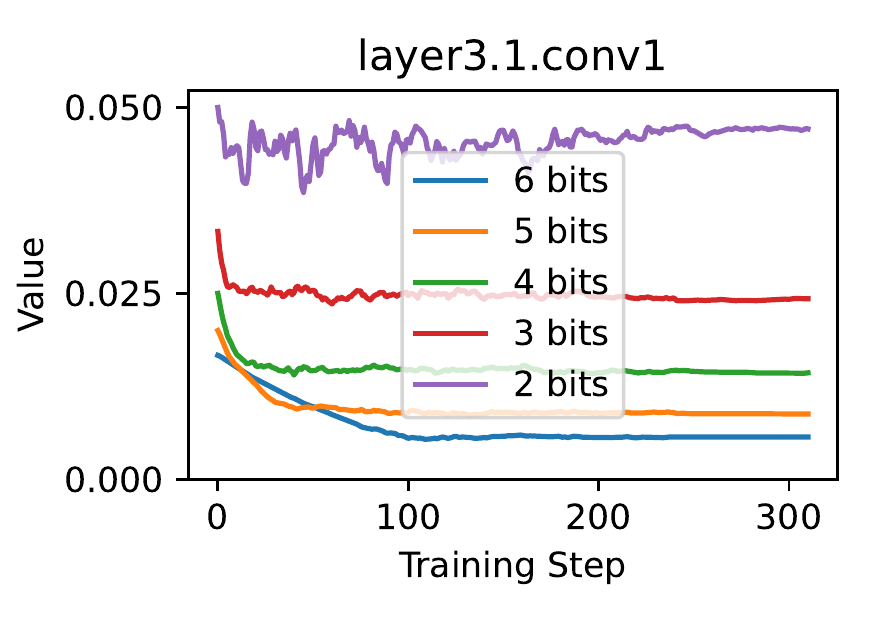}
\end{minipage}
\hspace{-0.2cm}
\begin{minipage}[t]{0.25\linewidth}
    \includegraphics[width = 1\linewidth]{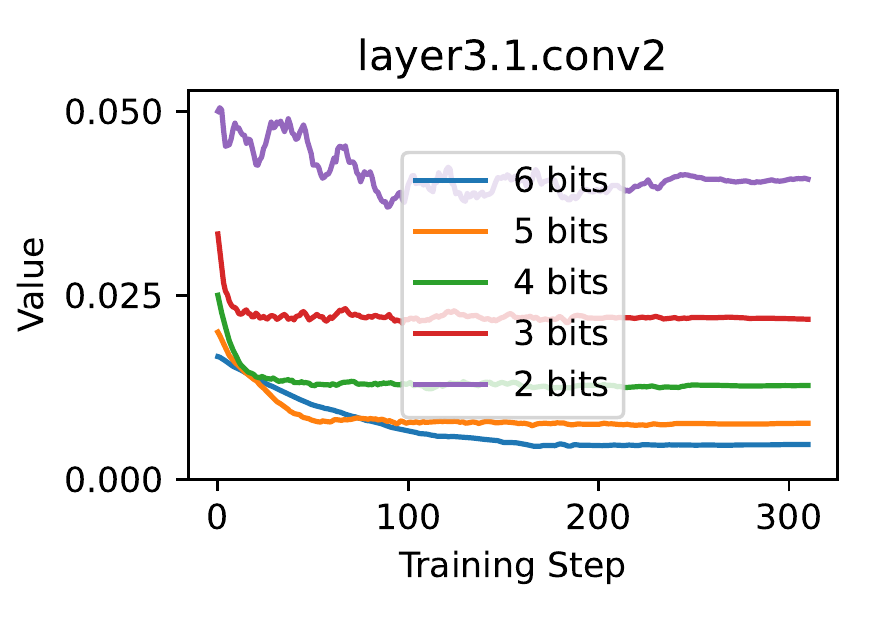}
\end{minipage}
\end{tabular}
\caption{The importance value of four layers for ResNet18.}
\label{fig:5epochs_resnet18_uniformly_initialization}
\end{figure}

As shown in Figure \ref{fig:5epochs_resnet18_uniformly_initialization}, after training of early instability, the scale factor still showed a significant difference at the end of training. 
That means the scale factor can still function consistently when using the same initialization for each layer. 
Nevertheless, we find that, compared to the same initialization value scheme, initialization with statistics \cite{bhalgat2020lsq+,esser2019learned} can speedup and stabilize the training process compared to the same initialization value strategy for each layers, thus we still use the statistics initialization scheme in our experiments.

\subsection{One-time Training for Importance Derivation}
\label{sec:one_time_importance_training}
Suppose the bit-width options for weights and activations are $\mathcal{B}=\{b_0, ..., b_{n-1}\}$, there are $M = 2 \times L \times n$ importance indicators for an $L$-layers network. Training these $M$ indicators separately requires $M$ training sessions, which induces huge extra training costs. 
Therefore, we propose a joint training scheme to obtain importance indicators of all layers corresponding $n$ bit-width options $\mathcal{B}$ at once training. 

Specifically, we use a bit-specific importance indicator instead of the original notion $s$ in Equation \ref{eq:preliminary} for each layer. That is, for the weights and activaions of layer $l$, we use the notion $s_{w, i}^{(l)}$ and $s_{a, j}^{(l)}$ as the importance indicator for $b_i \in \mathcal{B}$ of weights and $b_j \in \mathcal{B}$ of activations. In this way, $n$ different importance indicators can exist for each layer in a single training session. 
It is worth noting that the importance indicator parameters are only a tiny percentage of the overall network parameters, thus do not incur too much GPU memory overhead.
For example, for ResNet18, if there are 5 bit-width options per layer, we have $M=2\times19\times5=190$, while the whole network has more than 30 million parameters. 

At each training step $t$, we first perform $n$ times forward and backward propagation corresponding to $n$ bit-width options (\emph{i.e.,} respectively using same bit-width $b_k \in \mathcal{B}, k=0,..,n-1$ for each layer), and inspired by one-shot NAS \cite{guo2020single,chu2021fairnas} we then introduce one randomly bit-width assignment process for each layer to make sure different bit-widths in different layers can communicate with each other. 
We define the above procedure as an atomic operation of importance indicators update, in which only the gradients are calculated $n+1$ times, but the importance indicators are not updated during the execution of the operation. 
After that, we aggregate the above gradients and use them to update the importance indicators. 
See the Supplementary Material for details and pseudocode.

\begin{figure}[!htb]
\hspace{-0.3cm}
\subfigure[ResNet18]{
\begin{minipage}[t]{0.26\linewidth}
    \includegraphics[width = 1\linewidth]{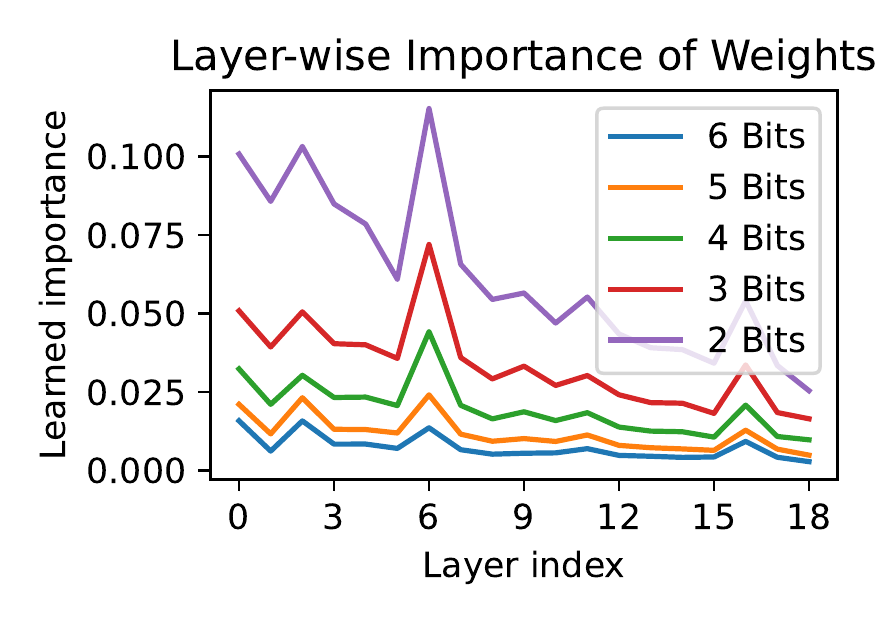}
\end{minipage}
\hspace{-0.3cm}
\begin{minipage}[t]{0.26\linewidth}
    \includegraphics[width = 1\linewidth]{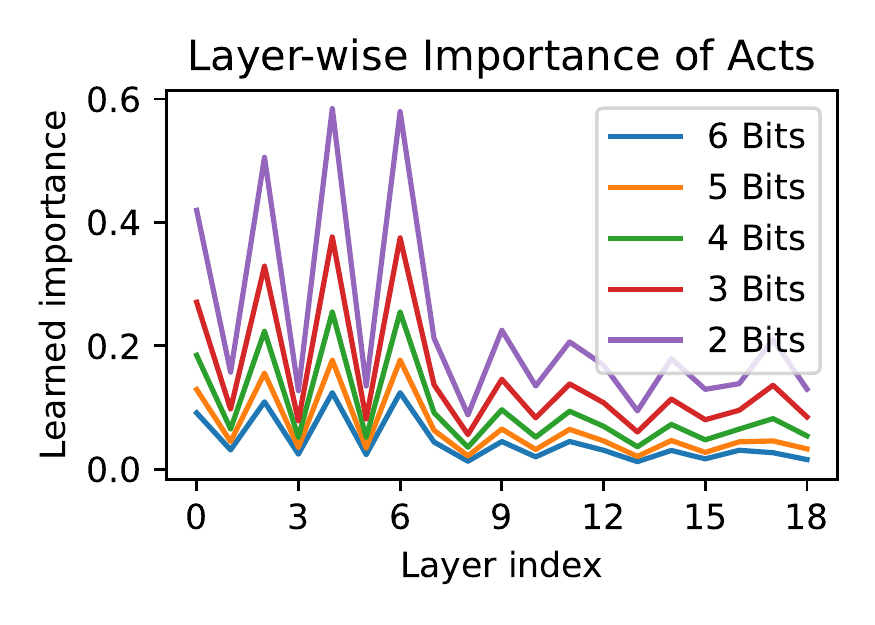}
\end{minipage}
}
\hspace{-0.4cm}
\subfigure[ResNet50]{
\begin{minipage}[t]{0.26\linewidth}
    \includegraphics[width = 1\linewidth]{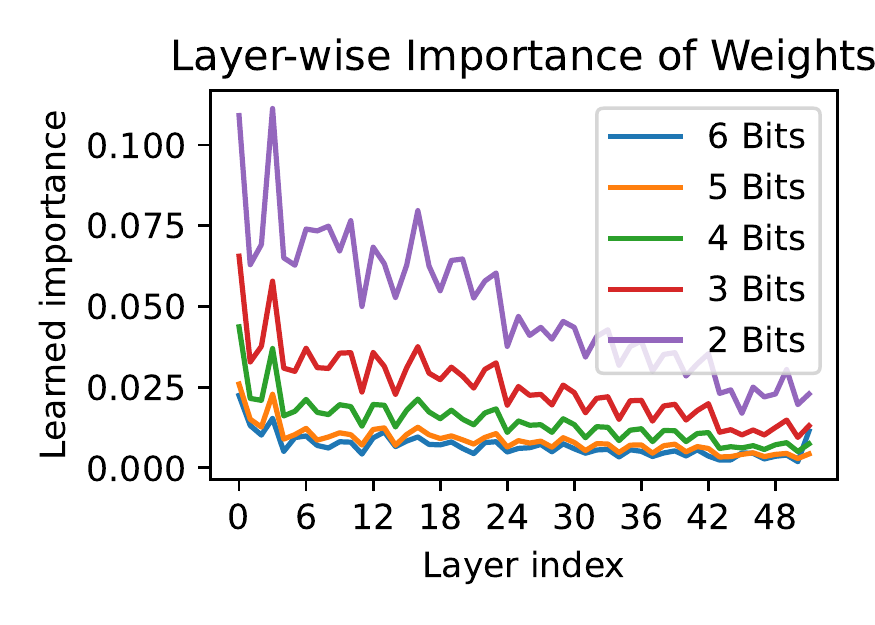}
\end{minipage}
\hspace{-0.3cm}
\begin{minipage}[t]{0.26\linewidth}
    \includegraphics[width = 1\linewidth]{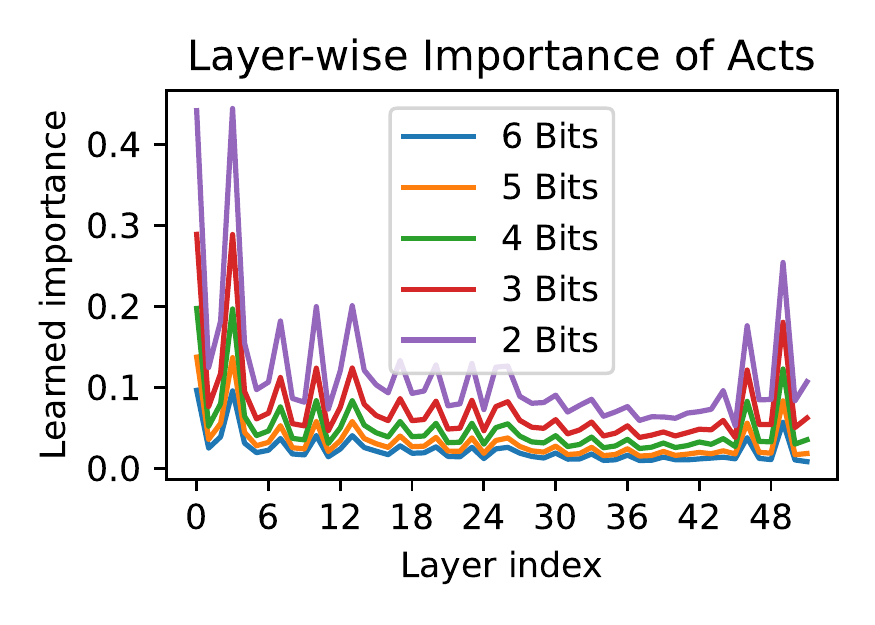}
\end{minipage}
}
\caption{The importance indicators for ResNet18 and ResNet50.}
\label{fig:layerwise_importance_for_resnet}
\end{figure}

We show in Figure \ref{fig:layerwise_importance_for_resnet} all the layer importance indicators obtained by this method in a single training session.
We observe that the top layers always show a higher importance value, indicating that these layers require higher bit-widths.

\subsection{Mixed-Precision Quantization Search Through Layer-wise Importance}
Now, we consider using these learned importance indicators to allocate bit-widths for each layer automatically. 
Since these indicators reflect the corresponding layer's contribution to final performance under certain bit-width, we no longer need to use iterative accuracy to evaluate the bit-width combination. 

As shown in Figure \ref{fig:mbv1_separate_importance}, the DW-convs always have a higher importance score than PW-convs, and the importance score rise when bit-width reduce, then DW-convs should be quantized to higher bit-width than PW-convs, \emph{e.g.,} 2 bits for PW-convs and 4 bits for DW-convs. 
For layer $l$, we use a binary variable $x_{i,j}^{(l)}$ representing the bit-width combination $(b^{(l)}_w, b^{(l)}_a)=(b_i, b_j)$ that $b_i$ bits for weights and $b_j$ bits for activations, whether it is selected or not. 
Under the given constraint $C$, our goal is to minimize the summed value of the importance indicator of every layer. 
Based on that, we reformulate the mixed-precision search into a simple ILP problem as Equation \ref{eq:search_ilp}:
\begin{subequations}\label{eq:search_ilp}
\begin{align}
\mathop{\arg\min}\limits_{\{x^{(l)}_{i,j}\}_{l=0}^L} \sum_{l=0}^L(s_{a, j}^{(l)}+\alpha \times s_{w, i}^{(l)}) \times x^{(l)}_{i,j}  \ \  \tag{\ref{eq:search_ilp}}
\end{align} 
\begin{alignat}{2}
\text{s.t.} 
\quad & \sum_i\sum_j x^{(l)}_{i,j}=1 \label{subeq:sumed_x} \\
& \sum_l \sum_i \sum_j BitOps\left(l, x^{(l)}_{i,j}\right) \leq C \label{subeq:bitops_c} \\
\text{vars} 
\quad & x^{(l)}_{i,j} \in \{0,1\}   
\end{alignat}
\end{subequations}
where Equation \ref{subeq:sumed_x} denotes only one bit-width combination selected for layer $l$, Equation \ref{subeq:bitops_c} denotes the summed BitOps of each layer constrains by $C$. Depending on the deployment scenarios, it can be replaced with other constraints, such as compression rate.
$\alpha$ is the hyper-parameter used to form a linear combination of weights and activations importance indicators.
Therefore, the final bit-width combination of the whole network $\mathcal{S^*}$ can be obtained by solving Equation \ref{eq:search_ilp}.

Please note that, since Equation \ref{eq:search_ilp} do not involve any training data, we no longer need to perform iterative evaluations on the training set as previous works. Thus the MPQ policy search time can be saved exponentially. 
We solve this ILP by a python library PuLP \cite{mitchell2011pulp}, elapsed time of the solver for ResNet18 is 0.06 seconds on an 8-core Apple M1 CPU. 
More details about MPQ policy search efficiency please refer \cref{sec:MPQ Policy Search Efficiency}.

\section{Experiments}
In this section, we conduct extensive experiments with the networks ResNet-18/50 \cite{he2016deep} and MobileNetv1 \cite{howard2017mobilenets} on ImageNet \cite{deng2009imagenet} classification. 
We compare our method with the fixed-precision quantization methods including 
PACT \cite{choi2018pact}, PROFIT \cite{park2020profit}, LQ-Net \cite{zhang2018lq}, 
and layer-wise MPQ methods HAQ \cite{wang2019haq}, AutoQ \cite{lou2019autoq}, SPOS \cite{guo2020single}, DNAS \cite{wu2018mixed}, BP-NAS \cite{yu2020search}, MPDNN \cite{uhlich2019mixed}, HAWQ \cite{dong2019hawq}, HAWQv2 \cite{dong2019hawq2}, DiffQ \cite{defossez2021differentiable} and MPQCO \cite{chen2021towards}.
Experimental setups can be found in the Supplementary Material.

\begin{table}[h]
\centering

\caption{Results for ResNet18 on ImageNet with BitOps constraints. ``W-bits'' and ``A-bits'' indicate bit-width of weights and activations respectively. ``MP'' means mixed-precision quantization. ``Top-1/Quant'' and ``Top-1/FP'' indicates the top-1 accuracy of quantized and \textbf{F}ull-\textbf{P}recision model. ``Top-1/Drop'' = ``Top-1/FP'' $-$ ``Top-1/Quant''.}
\setlength{\tabcolsep}{0.25mm}
\centering
\begin{tabular}{c|cccccc}
\hline

\quad Method \quad & \quad W-bits & \quad A-bits  & \quad Top-1/Quant  & \quad Top-1/FP  &  \quad Top-1/Drop  & \quad BitOps (G) \\ \hline

PACT    & 3     & 3     & 68.1 & 70.4   & -2.3          & 23.09  \\
LQ-Net  & 3     & 3     & 68.2 & 70.3   & -2.1          & 23.09  \\
Nice    & 3     & 3     & 67.7 & 69.8   & -2.1          & 23.09  \\
AutoQ   & 3MP   & 3MP   & 67.5 & 69.9   & -2.4          & -      \\
SPOS    & 3MP   & 3MP   & 69.4 & 70.9   & -1.5          & 21.92  \\
DNAS    & 3MP   & 3MP   & 68.7 & 71.0   & -2.3          & 25.38  \\
\hdashline
Ours    & 2.5MP & 3MP   & 68.7 & 69.6   & -0.9          & \textbf{19.81}  \\
Ours    & 3MP   & 3MP   & 69.0 & 69.6   & \textbf{-0.6}          & 23.07  \\
Ours    & 3MP   & 3MP   & \textbf{69.7} & 70.5   & -0.8          & 23.07  \\
\hline
PACT    & 4     & 4     & 69.2 & 70.4   & -1.2          & 33.07  \\
LQ-Net  & 4     & 4     & 69.3 & 70.3   & -1.0          & 33.07  \\
Nice    & 4     & 4     & 69.8 & 69.8   & 0          & 33.07  \\
SPOS    & 4MP   & 4MP   & 70.5 & 70.9   & -0.4          & 31.81  \\
MPDNN   & 4MP   & 4MP   & 70.0 & 70.2   & -0.2          & -      \\
AutoQ   & 4MP   & 4MP   & 68.2 & 69.9   & -1.7          & -      \\
DNAS    & 4MP   & 4MP   & 70.6 & 71.0   & -0.4          & 33.61  \\
MPQCO   & 4MP   & 4MP   & 69.7 & 69.8   & -0.1          & -      \\
\hdashline
Ours    & 4MP   & 4MP   & 70.1 & 69.6   & \textbf{0.5}           & 33.05  \\
Ours    & 4MP   & 4MP   & \textbf{70.8} & 70.5   & 0.3           & 33.05  \\

\hline
\end{tabular}
\label{tab:resnet18_result}
\end{table}

\vspace{-0.3cm}
\subsection{Mixed-Precision Quantization Performance Effectiveness}
\subsubsection{ResNet18}
In Table \ref{tab:resnet18_result}, we show the results of three BitOps (computation cost) constrained MPQ schemes, \emph{i.e.,} 2.5W3A of 19.81G BitOps, 3W3A of 23.07G BitOps and 4W4A of 33.05G BitOps.

Firstly, we observe that in 3-bits level (\emph{i.e.,} 23.07G BitOps) results. 
We achieve a \emph{least} absolute top-1 accuracy drop than all methods. 
Please note that the accuracy of our initialization full-precision (FP) model is only 69.6\%, which is about 1\% lower than some MPQ methods such as SPOS and DNAS. 
To make a fair comparison, we also provide a result initializing by a higher accuracy FP model (\emph{i.e.,} 70.5\%). 
At this time, the accuracy of the quantized model improves 0.7\% and reaches 69.7\%, which surpasses all existing methods, especially DNAS 1.0\% while DNAS uses a 71.0\% FP model as initialization. 
It is noteworthy that a 2.5W3A (\emph{i.e.,} 19.81G BitOps) result is provided to demonstrate that our method causes less accuracy drop even with a much strict BitOps constraint.

Secondly, in 4-bits level results (\emph{i.e.,} 33.05G BitOps), we also achieve a highest top-1 accuracy than prior arts whether it is fixed-precision quantization method or mixed-precision quantization method. 
A result initialized by a higher FP precision model is also provided for a fair comparison.

\subsubsection{ResNet50}
In Table \ref{tab:resnet50_result}, we show the results that not only perform a BitOps constrainted MPQ search but also set a model size constraint (\emph{i.e.,} 12.2 $\times$ compression rate). 
We can observe that our method achieves a much better performance than PACT, LQ-Net, DeepComp, and HAQ, under a much smaller model size (\emph{i.e.,} more than 9MB vs. 7.97MB).
In addition, the accuracy degradation of our method is smaller than the criterion-based methods HAWQ, HAWQv2 and MPQCO, which indicates that our quantization-aware search and unlimited search space is necessary for discovering a well performance MPQ policy.
\begin{table}[h]
\centering
\caption{Results for ResNet50 on ImageNet with BitOps and compression rate constraints. ``W-C'' means weight compression rate, the size of original full-precision model is 97.28 (MB). ``Size'' means quantized model size (MB). 
}
\setlength{\tabcolsep}{0.25mm}
\begin{tabular}{c|ccccccc}
\hline
Method  & W-bits    & A-bits     & \quad Top-1/Quant  & \quad Top-1/Full & \quad  Top-1/Drop.  & W-C            & Size (M)\\ \hline
PACT    & 3     & 3     & 75.3 & 76.9   & -1.6          & 10.67$\times$  & 9.17\\
LQ-Net  & 3     & 3     & 74.2 & 76.0   & -1.8          & 10.67$\times$  & 9.17\\
DeepComp& 3MP   & 8     & 75.1 & 76.2   &  -1.1         & 10.41$\times$  & 9.36 \\ 
HAQ     & 3MP   & 8     & 75.3 & 76.2   &  -0.9         & 10.57$\times$  & 9.22  \\
DiffQ   & MP    & 32    & 76.3 & 77.1   & -0.8          & 11.1$\times$   & 8.8    \\
BP-NAS  & 4MP   & 4MP   & 76.7 & 77.5   & -0.8          & 11.1$\times$   & 8.76     \\
AutoQ   & 4MP   & 3MP   & 72.5 & 74.8   & -2.3          & -              & -    \\
HAWQ    &  MP   &  MP   & 75.5 & 77.3   & -1.8          & 12.2$\times$   & \textbf{7.96}\\
HAWQv2  &  MP   &  MP   & 75.8 & 77.3   & -1.5          & 12.2$\times$   & 7.99\\
MPQCO   & 2MP   & 4MP   & 75.3 & 76.1   & -0.8          & 12.2$\times$   & 7.99\\
\hdashline
Ours    & 3MP   & 4MP   & \textbf{76.9} & 77.5   & \textbf{-0.6}         & \textbf{12.2$\times$}   & 7.97\\
\hline
\end{tabular}
\label{tab:resnet50_result}
\end{table}
\vspace{-1cm}
\subsubsection{MobileNetv1}
In Table \ref{tab:mobilenetv1_result}, we show the results of two BitOps constrainted including a 3-bits level (5.78G BitOps) and a 4-bits level (9.68G BitOps).
Especially in the 4-bit level result, we achieve a meaningful accuracy improvement (up to 4.39\%) compared to other MPQ methods.

In Table \ref{tab:mobilenetv1_weight_only_result}, we show the weight only quantization results.
We find that the accuracy of our 1.79MB model even surpasses that of the 2.12M HMQ model.

\begin{table*}[!htb]
\begin{minipage}[t]{6.1cm}
\centering
\footnotesize
\makeatletter\def\@captype{table}\makeatother\caption{
Results for MobileNetv1 on ImageNet with BitOps constraints. 
``W-b'' and ``A-b'' means weight and activation bit-widths. 
``Top-1'' and ``Top-5'' represent top-1 and top-5 accuracy of quantized model respectively. 
``B (G)'' means BitOps (G).}
\begin{tabular}{cccccc}
\hline
Method   & W-b    & A-b     & Top-1 & Top-5   & B (G)  \\ \hline
PROFIT   & 4    & 4     & 69.05    & 88.41      & 9.68  \\
PACT     & 6    & 4     & 67.51    & 87.84      & 14.13 \\
HMQ      & 3MP  & 4MP   & 69.30    & -          & -     \\
HAQ      & 4MP  & 4MP   & 67.45    & 87.85      & -     \\
HAQ      & 6MP  & 4MP   & 70.40    & 89.69      & -     \\

\hdashline
Ours    & 3MP  & 3MP   &  \textbf{69.48}   & \textbf{89.11} & \textbf{5.78}\\
Ours    & 4MP  & 4MP   &  \textbf{71.84}   & \textbf{90.38} & \textbf{9.68} \\ \hline
\label{tab:mobilenetv1_result}
\end{tabular}
\end{minipage}\hspace{2mm}
\begin{minipage}[t]{5.8cm}
\centering
\footnotesize
\makeatletter\def\@captype{table}\makeatother\caption{Weight only quantization results for MobileNetv1 on ImageNet.
``W-b'' means weight bit-widths. ``S (M)'' means quantized model size (MB).}
    \begin{tabular}{ccccc}
    
\hline
Method  & W-b & Top-1 & Top-5 & S (M)      \\ \hline
DeepComp& 3MP     & 65.93    & 86.85    & 1.60         \\
HAQ     & 3MP     & 67.66    & 88.21    & 1.58         \\
HMQ     & 3MP     & 69.88    & -        & \textbf{1.51}         \\
\hdashline
Ours    & 3MP     & \textbf{71.57}    & \textbf{90.30}    & 1.79         \\ \hline
PACT    & 8       & 70.82    & 89.85    & 4.01         \\
DeepComp& 4MP     & 71.14    & 89.84    & 2.10         \\
HAQ     & 4MP     & 71.74    & 90.36    & \textbf{2.07}         \\
HMQ     & 4MP     & 70.91    & -        & 2.12         \\
\hdashline
Ours    & 4MP     & \textbf{72.60}    & \textbf{90.83} & 2.08\\ \hline
\label{tab:mobilenetv1_weight_only_result}
\end{tabular}
\vspace{-1.2cm}
\end{minipage}
\end{table*}
\subsection{Mixed-Precision Quantization Policy Search Efficiency}
\label{sec:MPQ Policy Search Efficiency}
Here, we compare the efficiency of our method to other SOTAs MPQ algorithms with unlimited search space (\emph{i.e.,} MPQ for both weights and activations instead of weights only MPQ, layer-wise MPQ instead of block-wise).
Additional results about DiffQ \cite{defossez2021differentiable} and HAWQ \cite{dong2019hawq} can be found in the Supplementary Material.

The time consumption of our method consists of 3 parts. Namely,  
\emph{1)} Importance indicators training.
\emph{2)} MPQ policy search.
\emph{3)} Quantized model fine-tuning.
The last part is necessary for all MPQ algorithms while searching the MPQ policy is the biggest bottleneck (\emph{e.g.,} AutoQ needs more than 1000 GPU-hours to determine the final MPQ policy), thus we mainly focus on the first two parts.

\subsubsection{Comparison with SOTAs on ResNet50} 
The time consumption of the first part is to leverage the joint training technique (see \cref{sec:one_time_importance_training}) to get importance indicators for all layers and their corresponding bit-widths, but it only needs to be done once. 
It needs to train the network about 50 minutes (using 50\% data of training set) on 4 NVIDIA A100 GPUs (\emph{i.e.,} 3.3 GPU-hours). 
The time consumption of the second part is to solve the ILP problem. It consumes 0.35 seconds on a six-core Intel i7-8700 (at 3.2 GHz) CPU, which is negligible. 

Hence, suppose we have different $z$ devices with diverse computing capabilities to deploy, our method consumes $50+0.35 \times \frac{1}{60} \times z$ minutes to finish the whole MPQ search processes. 

\textbf{Compared with the search-based approach} AutoQ \cite{lou2019autoq} needs 1000 GPU-hours to find the MPQ policy for a single device, which means it needs $1000z$ GPU-hours to search MPQ policies for these $z$ devices. 
Thus we achieve about $\mathbf{330z\times}$ speedup and obtain a higher accuracy model simultaneously. 

\textbf{Compared with the criterion-based approach}, 
HAWQv2 \cite{dong2019hawq2} takes 30 minutes on 4 GPUs to approximate the Hessian trace. 
The total time consumption of HAWQv2 for these $z$ devices is $30+c \times \frac{1}{60} \times z$ minutes, and $c$ is the time consumption for solving a Pareto frontier based MPQ search algorithm with less than 1 minute.
Thus if $z$ is large enough, our method has almost the same time overhead as HAWQv2.
If $z$ is small, \emph{e.g.,} $z=1$, our method only needs a one-time additional 20-minute investment for the cold start of first part, but resulting in a significant accurate model (\emph{i.e.,} 1.1\% top-1 accuracy improvement).

\subsection{Ablation Study}
In previous analysis, we empirically verify that the layers with bigger scale factor values are more sensitive to quantization when their quantization bit-width is reduced.
Motivated by this, we propose our ILP-based MPQ policy search method. 
However, an intuitive question is \emph{what if we reverse the correlation between scale factors and sensitivity}. 
Namely, what if we gave the layers with smaller scale factor values more bit-widths instead of fewer bit-widths. 
And, what if we gave the layers with bigger scale factor values fewer bit-widths instead of more bit-widths. 

The result is shown in Table \ref{tab:ablation_study}, we use ``Ours-R'' to denote the result of reversed bit-width assignment manner; ``Ours'' results come from Table \ref{tab:mobilenetv1_result} directly to represent the routine (not reversed) bit-width assignment manner. 

We observe that ``Ours-R'' has 6.59\% top-1 accuracy lower than our routine method under the same BitOps constraint. 
More seriously, it has 4.23\% absolute accuracy gap between ``Ours-R'' (with 4-bits level constrainted, \emph{i.e.,} 9.68 BitOps) and a 3-bits level (\emph{i.e.,} 5.78G BitOps) routine result.
Such a colossal accuracy gap demonstrates that our ILP-based MPQ policy search method is reasonable.

\begin{table}[h]
\centering
\caption{Ablation study for MobileNetv1 on ImageNet.}
\begin{tabular}{c|ccccc}
\hline
Method    & W-bits & A-bits    & Top-1/Quant  & Top-5/Quant & BitOps \\ \hline
Ours & 3MP & 3MP & 69.48        & 89.11      & 5.78 \\
Ours & 4MP & 4MP & 71.84        & 90.38      & 9.68 \\
Ours-R & 4MP & 4MP & 65.25        & 86.15      & 9.68   \\

\hline
\end{tabular}
\label{tab:ablation_study}
\vspace{-10pt}
\end{table}
\section{Conclusion}
In this paper, we propose a novel MPQ method that leverages the unique parameters in quantization, namely the scale factors in the quantizer, as the importance indicators to assign the bit-width for each layer. 
We demonstrate the association between these importance indicators and the quantization sensitivity of layers empirically. 
We conduct extensive experiments to verify the effectiveness of using these learned importance indicators to represent the contribution of certain layers under specific bit-width to the final performance, as well as to demonstrate the rationality of the bit-width assignment obtained by our method. 
\subsection*{Acknowledgements}
This work is supported in part by NSFC (Grant No. 61872215 and No. 62072440), the Beijing Natural Science Foundation (Grant No. 4202072), and Shenzhen Science and Technology Program (Grant No. RCYX20200714114523079). 
Yifei Zhu’s work is supported by SJTU Explore-X grant.

\clearpage
%
%
\bibliographystyle{splncs04}
\bibliography{egbib}

\begin{thebibliography}{10}
\providecommand{\url}[1]{\texttt{#1}}
\providecommand{\urlprefix}{URL }
\providecommand{\doi}[1]{https://doi.org/#1}

\bibitem{baskin2021nice}
Baskin, C., Zheltonozhkii, E., Rozen, T., Liss, N., Chai, Y., Schwartz, E.,
  Giryes, R., Bronstein, A.M., Mendelson, A.: Nice: Noise injection and
  clamping estimation for neural network quantization. Mathematics
  \textbf{9}(17), ~2144 (2021)

\bibitem{bhalgat2020lsq+}
Bhalgat, Y., Lee, J., Nagel, M., Blankevoort, T., Kwak, N.: Lsq+: Improving
  low-bit quantization through learnable offsets and better initialization. In:
  Proceedings of the IEEE/CVF Conference on Computer Vision and Pattern
  Recognition Workshops. pp. 696--697 (2020)

\bibitem{cai2017deep}
Cai, Z., He, X., Sun, J., Vasconcelos, N.: Deep learning with low precision by
  half-wave gaussian quantization. In: Proceedings of the IEEE conference on
  computer vision and pattern recognition. pp. 5918--5926 (2017)

\bibitem{cai2020rethinking}
Cai, Z., Vasconcelos, N.: Rethinking differentiable search for mixed-precision
  neural networks. In: Proceedings of the IEEE/CVF Conference on Computer
  Vision and Pattern Recognition. pp. 2349--2358 (2020)

\bibitem{chen2021bn}
Chen, B., Li, P., Li, B., Lin, C., Li, C., Sun, M., Yan, J., Ouyang, W.:
  Bn-nas: Neural architecture search with batch normalization. In: Proceedings
  of the IEEE/CVF International Conference on Computer Vision. pp. 307--316
  (2021)

\bibitem{chen2021towards}
Chen, W., Wang, P., Cheng, J.: Towards mixed-precision quantization of neural
  networks via constrained optimization. In: Proceedings of the IEEE/CVF
  International Conference on Computer Vision. pp. 5350--5359 (2021)

\bibitem{choi2018pact}
Choi, J., Wang, Z., Venkataramani, S., Chuang, P.I.J., Srinivasan, V.,
  Gopalakrishnan, K.: Pact: Parameterized clipping activation for quantized
  neural networks. arXiv preprint arXiv:1805.06085  (2018)

\bibitem{chu2021fairnas}
Chu, X., Zhang, B., Xu, R.: Fairnas: Rethinking evaluation fairness of weight
  sharing neural architecture search. In: Proceedings of the IEEE/CVF
  International Conference on Computer Vision. pp. 12239--12248 (2021)

\bibitem{defossez2021differentiable}
D{\'e}fossez, A., Adi, Y., Synnaeve, G.: Differentiable model compression via
  pseudo quantization noise. arXiv preprint arXiv:2104.09987  (2021)

\bibitem{deng2009imagenet}
Deng, J., Dong, W., Socher, R., Li, L.J., Li, K., Fei-Fei, L.: Imagenet: A
  large-scale hierarchical image database. In: 2009 IEEE conference on computer
  vision and pattern recognition. pp. 248--255. Ieee (2009)

\bibitem{dong2019hawq2}
Dong, Z., Yao, Z., Cai, Y., Arfeen, D., Gholami, A., Mahoney, M.W., Keutzer,
  K.: Hawq-v2: Hessian aware trace-weighted quantization of neural networks.
  In: Advances in neural information processing systems (2020)

\bibitem{dong2019hawq}
Dong, Z., Yao, Z., Gholami, A., Mahoney, M.W., Keutzer, K.: Hawq: Hessian aware
  quantization of neural networks with mixed-precision. In: Proceedings of the
  IEEE/CVF International Conference on Computer Vision. pp. 293--302 (2019)

\bibitem{esser2019learned}
Esser, S.K., McKinstry, J.L., Bablani, D., Appuswamy, R., Modha, D.S.: Learned
  step size quantization. In: International Conference on Learning
  Representations (2020)

\bibitem{guo2020single}
Guo, Z., Zhang, X., Mu, H., Heng, W., Liu, Z., Wei, Y., Sun, J.: Single path
  one-shot neural architecture search with uniform sampling. In: European
  Conference on Computer Vision. pp. 544--560. Springer (2020)

\bibitem{habi2020hmq}
Habi, H.V., Jennings, R.H., Netzer, A.: Hmq: Hardware friendly mixed precision
  quantization block for cnns. In: Computer Vision--ECCV 2020: 16th European
  Conference, Glasgow, UK, August 23--28, 2020, Proceedings, Part XXVI 16. pp.
  448--463. Springer (2020)

\bibitem{he2016deep}
He, K., Zhang, X., Ren, S., Sun, J.: Deep residual learning for image
  recognition. In: IEEE Conference on Computer Vision and Pattern Recognition
  (CVPR). pp. 770--778 (2016)

\bibitem{howard2017mobilenets}
Howard, A.G., Zhu, M., Chen, B., Kalenichenko, D., Wang, W., Weyand, T.,
  Andreetto, M., Adam, H.: Mobilenets: Efficient convolutional neural networks
  for mobile vision applications. arXiv preprint arXiv:1704.04861  (2017)

\bibitem{jung2019learning}
Jung, S., Son, C., Lee, S., Son, J., Han, J.J., Kwak, Y., Hwang, S.J., Choi,
  C.: Learning to quantize deep networks by optimizing quantization intervals
  with task loss. In: Proceedings of the IEEE/CVF Conference on Computer Vision
  and Pattern Recognition. pp. 4350--4359 (2019)

\bibitem{lecun1990optimal}
LeCun, Y., Denker, J.S., Solla, S.A.: Optimal brain damage. In: Advances in
  neural information processing systems. pp. 598--605 (1990)

\bibitem{liu2017learning}
Liu, Z., Li, J., Shen, Z., Huang, G., Yan, S., Zhang, C.: Learning efficient
  convolutional networks through network slimming. In: Proceedings of the IEEE
  international conference on computer vision. pp. 2736--2744 (2017)

\bibitem{lou2019autoq}
Lou, Q., Guo, F., Kim, M., Liu, L., Jiang, L.: Autoq: Automated kernel-wise
  neural network quantization. In: International Conference on Learning
  Representations (2020)

\bibitem{mitchell2011pulp}
Mitchell, S., OSullivan, M., Dunning, I.: Pulp: a linear programming toolkit
  for python. The University of Auckland, Auckland, New Zealand p.~65 (2011)

\bibitem{park2020profit}
Park, E., Yoo, S.: Profit: A novel training method for sub-4-bit mobilenet
  models. In: European Conference on Computer Vision. pp. 430--446. Springer
  (2020)

\bibitem{uhlich2019mixed}
Uhlich, S., Mauch, L., Cardinaux, F., Yoshiyama, K., Garcia, J.A., Tiedemann,
  S., Kemp, T., Nakamura, A.: Mixed precision dnns: All you need is a good
  parametrization. In: International Conference on Learning Representations
  (2020)

\bibitem{wang2019haq}
Wang, K., Liu, Z., Lin, Y., Lin, J., Han, S.: Haq: Hardware-aware automated
  quantization with mixed precision. In: Proceedings of the IEEE/CVF Conference
  on Computer Vision and Pattern Recognition. pp. 8612--8620 (2019)

\bibitem{wu2018mixed}
Wu, B., Wang, Y., Zhang, P., Tian, Y., Vajda, P., Keutzer, K.: Mixed precision
  quantization of convnets via differentiable neural architecture search. arXiv
  preprint arXiv:1812.00090  (2018)

\bibitem{yu2020search}
Yu, H., Han, Q., Li, J., Shi, J., Cheng, G., Fan, B.: Search what you want:
  Barrier panelty nas for mixed precision quantization. In: European Conference
  on Computer Vision. pp. 1--16. Springer (2020)

\bibitem{zhang2018lq}
Zhang, D., Yang, J., Ye, D., Hua, G.: Lq-nets: Learned quantization for highly
  accurate and compact deep neural networks. In: Proceedings of the European
  conference on computer vision (ECCV). pp. 365--382 (2018)

\bibitem{zhou2017incremental}
Zhou, A., Yao, A., Guo, Y., Xu, L., Chen, Y.: Incremental network quantization:
  Towards lossless cnns with low-precision weights. arXiv preprint
  arXiv:1702.03044  (2017)

\bibitem{zhou2016dorefa}
Zhou, S., Wu, Y., Ni, Z., Zhou, X., Wen, H., Zou, Y.: Dorefa-net: Training low
  bitwidth convolutional neural networks with low bitwidth gradients. arXiv
  preprint arXiv:1606.06160  (2016)

\end{thebibliography}
\end{document}